\documentclass[11pt]{article}
\usepackage{acl2015}
\usepackage{times}
\usepackage{url}
\usepackage{latexsym}
\usepackage{graphicx}
\usepackage{amsmath}

\DeclareMathOperator*{\argmax}{arg\,max}

\title{Synthesizing Novel Pairs of Image and Text}

\author{Jason Xie \\
  {\tt jxieeducation@berkeley.edu} \\\And 
  Tingwen Bao \\
  {\tt tingwenbao@berkeley.edu} \\}

\begin{document}
\maketitle
\begin{abstract}
	Generating novel pairs of image and text is a problem that combines computer vision and natural language processing. In this paper, we present strategies for generating novel image and caption pairs based on existing captioning datasets. The model takes advantage of recent advances in generative adversarial networks and sequence-to-sequence modeling. We make generalizations to generate paired samples from multiple domains. Furthermore, we study cycles -- generating from image to text then back to image and vise versa, as well as its connection with autoencoders.
\end{abstract}

\section{Introduction}

The scarcity in the availability of image captioned datasets raises the question of whether it'd be feasible to synthesize artificial samples that contain high quality pairs of image and text. Such an application could help improve the performance of neural networks on tasks such as image captioning and image classification. Commercial applications also include content creation, drug discovery and news entertainment. 

Recent advances in generative adversarial networks (GAN) and sequence to sequence modeling have made image to text and text to image feasible. Our work combines these two paradigms and explore the implications of combining the two models.

Our main accomplishments include: 
1) To the best of our knowledge, the first study on synthesizing novel pairs of image and text
2) An analysis of cycles, from image to text back to image, and a comparison with autoencoders.

\begin{figure} [h!]
  \centering
	\includegraphics[width=0.45\textwidth]{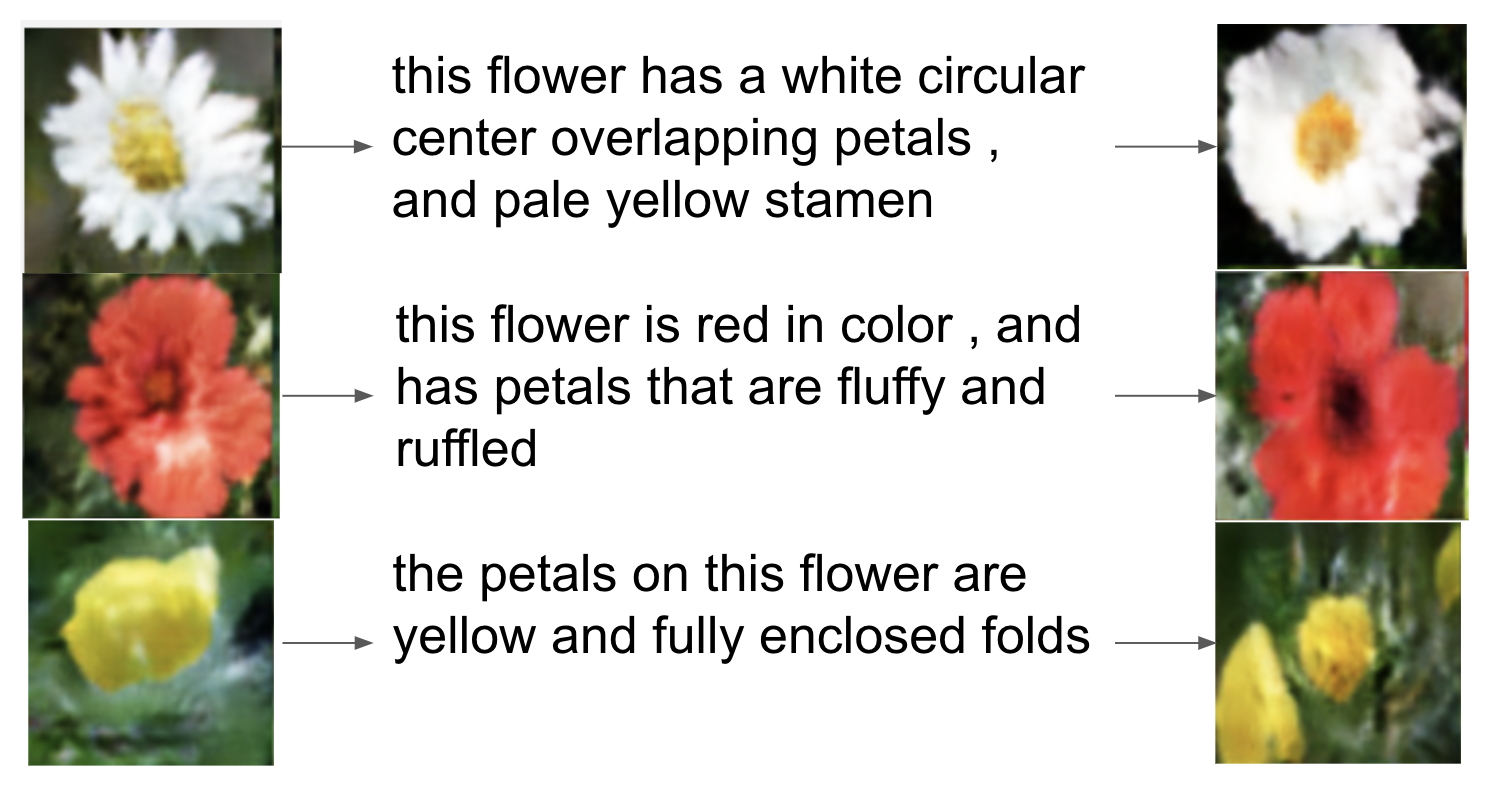}
  \caption{Example of going from image to text then back to image. Note that the content of the image remains the same.}
\end{figure}

\section{Background}

\subsection{Generative Adversarial Networks}

Generative image modeling has seen tremendous improvements in the past few years with the emergence of deep learning techniques. Generative Adversarial Networks pioneered by Goodfellow {\em et al.} has since opened doors to image generation with deep-convolution generative models (DCGAN)  ~\cite{Goodfellow:14}. In the paper "Generative Adversarial Text to Image Synthesis", Reed {\em et al.} provided a multi-stage approach for generating image from text  ~\cite{Reed:16}. The authors first learned a sequence encoder that learns the discriminative text feature representations. Then the authors trained a deep-convolution generative model (DCGAN) conditioned on the text embeddings. This end to end approach provided a performant way to generate images based on text. 

\subsection{Image Captioning}

On the flip side of text to image, the problem of generating natural language descriptions from images has seen tremendous improvements due to the clever amalgamation of deep convolution networks with recurrent nerual networks. Techniques such as seq2seq by Sutskever {\em et al.} has shown great potential in natural language processing tasks such as machine translation ~\cite{Sutskever:14}. In "Show and Tell: A Neural Image Caption Generator", Vinyals {\em et al.}  demonstrated that descriptive and coherent captions can be generated by feeding the last convolutional activations from an image classification network through a long short term memory (LSTM) network ~\cite{Vinyals:15}. 

\subsection{Dataset Augmentation}

Inspired by the above advances, we try to tackle the problem of dataset augmentation through synthesizing novel image and text pairs. There has been a plethora of work detailing ways of creating additional samples in the imbalanced learning literature, where new examples are created for the minority class in order to prevent overfitting. In Synthetic Minority Over-sampling Technique (SMOTE), examples are created in feature-space by randomly selecting pairs from the original dataset ~\cite{Chawla:02}. In Adasyn, a new sample is generated by combining an existing sample with weighted differences of the sample with its neighbors. In Parsimonious Mixture of Gaussian Trees, new samples are created by first fitting the existing minority class with mixture of gaussians, and then sampling from them ~\cite{Cao:14}. We take inspiration from much of the literature in imbalanced learning, where novel samples are generated for the minority class in order to improve learning.

In section 3 of the paper, we cover how we are re-purposing Reed {\em et al.}'s GAN-CLS to generate new image and text pairs, as well as its connection with autoencoders. In section 4 of the paper, we show qualitative results and an analysis of our approach. 

\section{Model}

In this paper, we utilize a neural framework to generate images and text. Recent advances in generative adversarial network has shown the feasibility of translating image to text, $V = F(T)$, as well as translating text to image, $T = G(V)$. Combining the two, we gain the ability to generate novel image text pairs. 

Specifically in terms of image and text, our problem is formulated as generating novel images $v_{new}$ and $t_{new}$ based on existing pairs of images and texts. We formulate our challenge as $(v_{new}, t_{new})$ \textasciitilde $S(V, T)$, where $(v_{new}, t_{new}) \notin {v_{i..N}, t_{i..N}}$, the existing dataset. 

We approach the paired generation problem by using a 2-stepped approach:
\begin{enumerate}
\item Source domain generation: sample from a source domain: $v_i$ \textasciitilde $V$
\item Target domain generation: generate the target domain conditioned on the source domain, $v_i$, $t_i$ \textasciitilde $P(T | v_i)$
\end{enumerate}

In the case where the source domain is text, we first sample a text, then generate an image based on the text sample; and vise versa, we first sample an image, then generate a text caption based on the image sample. 

\subsection{Source domain generation}

The source domain generation problem is a problem of given examples $v_{1..N}$, generate novel samples $v_{new}$. This problem requires us to construct novel examples, instead of sampling the existing dataset. Because of the high dimensionality of the image and text domain, we instead generate novel embeddings instead. The encoding of text and image are learned during the training phase of GAN-GLS. $\phi(v_i)$ is the last convolutional activation in the discriminator, and $\psi(t_i)$ is the encoder that the generator is conditioned on. In order to construct $\phi(v_i)_{new}$ and $\psi(t_i)_{new}$, We propose 2 approaches inspired by the existing imbalance class learning literature, prototype-based and density-based.

\begin{figure} [h!]
  \centering
	\includegraphics[width=0.2\textwidth]{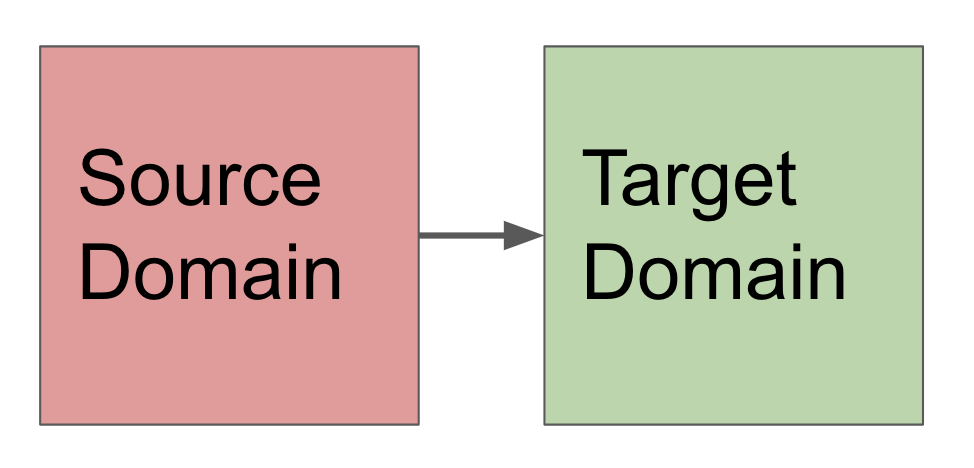}
  \caption{2 step approach to image and text pair generation. 1) generate a novel sample from the source domain. 2) translate the sample from the source domain to the target domain.}
\end{figure}

\subsubsection{Prototype based}

In the first method, we make modifications to the existing samples, prototypes. Prototypes $v_i$ \textasciitilde $V$ are examples from the dataset. Our contribution draws heavily from techniques such as SMOTE that synthesize new examples based on pairs of existing examples.

A new image embedding can be produced via $$\lambda \phi(v_i) + (1 - \lambda) \phi(v_j)$$ and vise versa for text $$\lambda \psi(t_i) + (1 - \lambda) \psi(t_j)$$. A major advantage of using existing prototypes is that the prototypes can be easily identified. The prototype method can be further extended by sampling $\phi(v_i)$ and $\phi(v_j)$ conditioned on additional information, such as the source class. 

\subsubsection{Density based}

In density-based source domain generation, we learn a generative model that sample from the distribution of the source domain. Then we sample from the learned distribution to obtain additional examples in the source domain. $\phi(v_i)$ \textasciitilde $\phi(V)$ 

In our application, we fit mixture models for both text and image embeddings respectively. $p(x) = \sum_{i=1}^K \omega_i N (x | \mu_i, \Sigma_i)$ where there are K mixtures, each parameterized by a Gaussian. 

\subsection{Target domain generation}

After generating the source domain, we want to generate the target domain conditioned on the source domain. This leads us to our formulation to go from text to image, $v_i = f(\psi(t_i))$, and image to text, $t_i = g(\phi(v_i))$. 

\subsubsection{Text to Image}

Like Reed {\em et al.} ~\cite{Reed:16}, we model text to image synthesis as a conditional generative adversarial network (cGAN). The generator network is denoted $$G : N(0, 1) \times \psi(t_i) \rightarrow v_i$$ where $N(0,1) \in R^Z$, $\psi(t_i) \in R^T$ and $v_i \in R^V$. On the other hand, the discriminator network is denoted $$D : v_i \times \psi(t_i) \rightarrow \{0, 1\}$$

The loss function is adopted from GAN-CLS, 
$$L_D \leftarrow log(D(v, t)) + (log (1 - D(v, t_{fake}))$$ $$ + log(1 - D(v_{fake}, t)) ) / 2$$ 
The intuition behind the loss is that it forces the discriminator to learn to discern: generated images with right captions $D(v, t)$, real images with wrong captions $D(v, t_{fake})$, and wrong images with the right captions $D(v_{fake}, t)$. 

\subsubsection{Image to Text}

Going from image to text, we train our model to maximize the likelihood of the correct caption. 

$$\theta^* = \argmax\limits_{\theta} \sum_{(v_i, t_i)} \log p(t_i | v_i ; \theta)$$

We estimate $p(t_i | v_i ; \theta)$ using a LSTM. Unlike Vinyals {\em et al.} ~\cite{Vinyals:15}, instead of training a new convolutional neural network, we take advantage of the discriminator trained in our conditional GAN from section 3.2.1 by utilizing the last convolutional layer of the CNN, $\phi(v)$.

$$p_{t_i} = LSTM(\phi(v_i))$$

Note that from the GAN-CLS architecture, $\phi(v_i)$ has not been concatenated with textual information $\psi(t_i)$. This means that $\phi(v_i)$ contains the information needed to distinguish real images with wrong captions and wrong images with the right captions. This leads us to conclude that $\phi(v_i)$ has the capability of generating descriptive captions. 

\subsection{Formulating a cycle: A to B to A }

\begin{figure} [h!]
  \centering
	\includegraphics[width=0.45\textwidth]{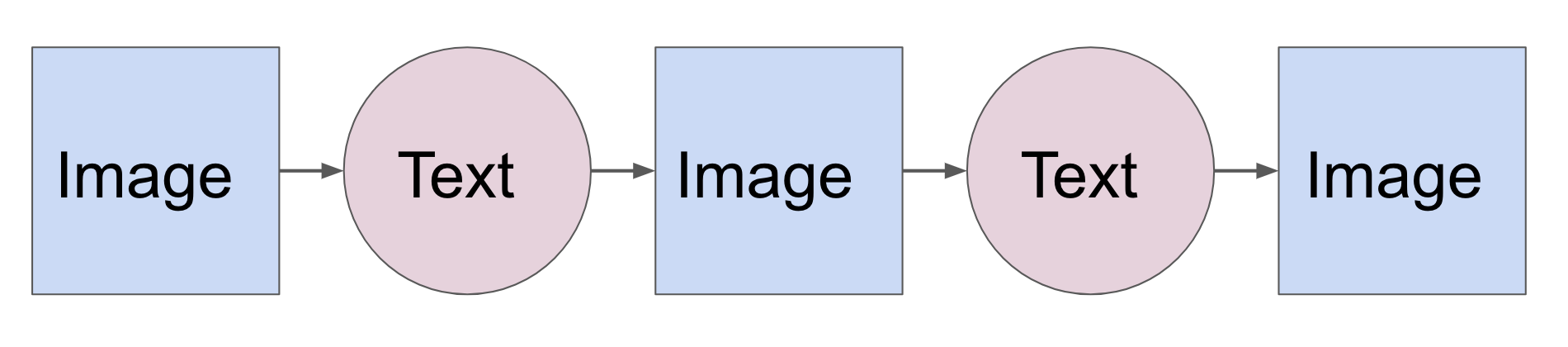}
  \caption{Going from one domain to another then back.}
\end{figure}

Combining $v_i = f(\psi(t_i))$, generating an image based on text, and $t_i = g(\phi(v_i))$, generating a text based on image, we study the implications of going from image to text then back to image, or vise versa, which we term cycles:

$$v_i^{'} = f(\psi(g(\phi(v_i))))$$

$$t_i^{'} = g(\phi(f(\psi(t_i))))$$

\subsubsection{Relation to Autoencoders}

The traditional autoencoder tries to minimize the reconstruction loss between its input and the reconstructed input. $J(W) = \sum_n \| x_i - \hat{x_i} \|$ Although not directly minimizing the loss, our formulation of cycles do have close ties with autoencoders.

As noted by Goodfellow {\em et al.}, GANs, in particular, G, is learning the true distribution of the input data. $p_g(v) = p_{data}(v)$. Conditional GAN, by extension, is also learning the true conditional distribution. $p_g(v | t) = p_{data} (v | t)$  This means that when going from text to image, the conditional distribution of the data is preserved.

On the other hand, when going from image to text, we maximize $p_{data}(t | v)$, which also means that the conditional distribution of the data is preserved.

However, it's important to note that $p(v | t)$ and $p(t | v)$ only preserve information that is shared between the image and text. Information not shared between the image and text are lost in the generation process. 

Therefore, we should expect the information loss within the embedding space, information shared between the two domains, to be minimal. This means that $cos (\phi(v), \phi(v'))$ should be close to $1$, where $v$ and $v'$ are the original and reconstructed images. However, we should not expect $cos (v, v')$ to also be similar, since $f$ and $g$ maximize the probability of the data conditioned on the shared information between image and text. 

\section{Result}

We experimented with the Oxford 102 flowers dataset, which consists of 102 classes of flowers, 8000 images along with 10 captions per each image. Using our model, we are able to successfully generate novel pairs of image and text.

\subsection{Source domain generation}

In {\em Figure 4, 5 and 6}, the source domain is text and the target domain is images. In {\em Figure 4}, we interpolate two text embeddings $\psi(t_i)$ and $\psi(t_j)$, with $\lambda$ linearly interpolating between $0$ and $1$. This demonstrates our ability to construct novel images based on text.

\begin{figure} [h!]
  \centering
	\includegraphics[width=0.45\textwidth]{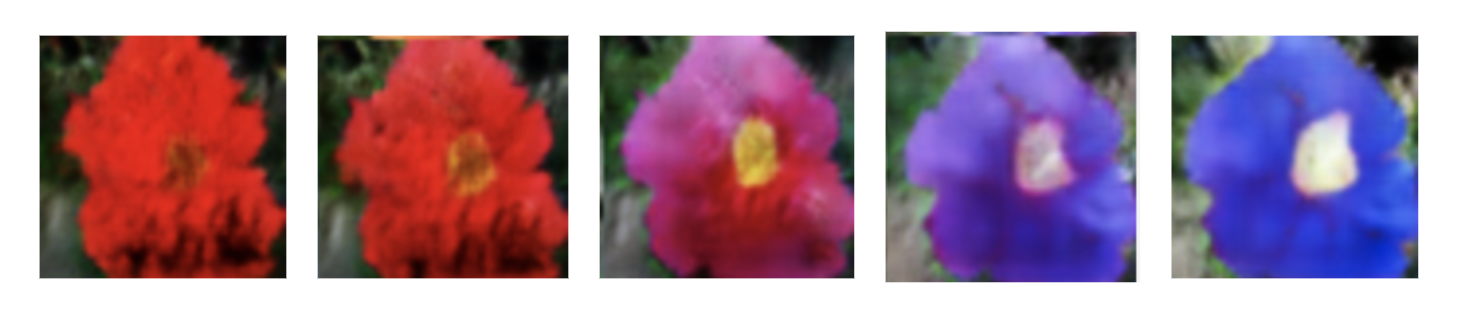}
  \caption{Combining two captions to create a new image. An example of prototype based source domain generation. The leftmost image is generated based on the caption "The flower is red" and the rightmost image is generated based on the caption "The flower is blue". The images in between are linear interpolations of the text embedding $\lambda \psi(t_1) + (1 - \lambda) \psi(t_2)$. }
\end{figure}

In {\em Figure 5}, we demonstrate the reverse result of {\em Figure 4} and combine two prototype images to generate novel text.

\begin{figure} [h!]
  \centering
	\includegraphics[width=0.45\textwidth]{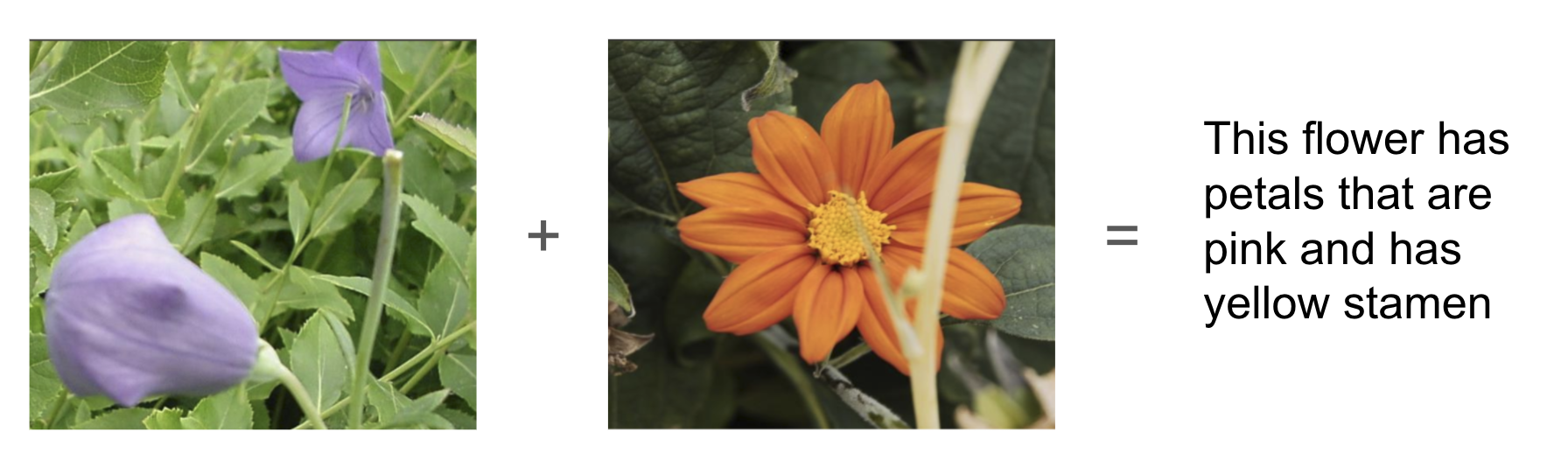}
  \caption{Combining two images to create a new caption. Another example of prototype based source domain generation. The caption on the right is generated by combining the embeddings of the images on the left $0.5 \phi(v_1) + 0.5 \phi(v_2)$.}
\end{figure}

In {\em Figure 6}, we explore density based source domain generation. Each image is generated by sampling from the same cluster in the Gaussian mixture model. 

\begin{figure} [h!]
  \centering
	\includegraphics[width=0.45\textwidth]{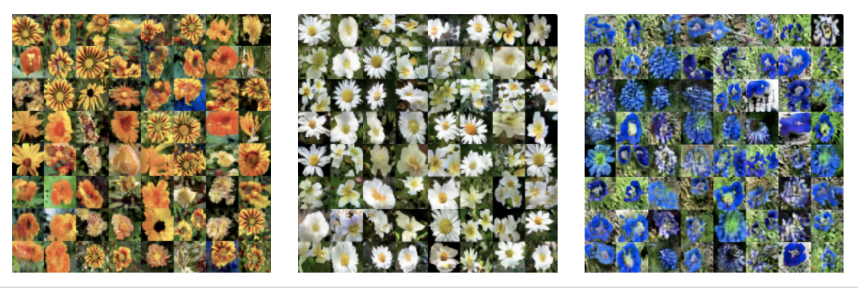}
  \caption{Creating flower images by sampling from the density model learned on the text embedding. Example of density based source domain generation. Each image contains examples sampled from the same cluster in the Gaussian mixture model. Note the homogeneity of images within the cluster.}
\end{figure}

\subsection{Cycle generation: A to B to A}

{\em Figure 1.} in section 1 shows an example of the cycle of of image to text to image. Note that the caption and image preserve the semantic attributes of the flowers instead of the raw pixel values. On the other hand, for text to image to text, {\em Figure 7.} below shows similar behavior, the meaning of the text stays the same, but not in exact words.

\begin{figure} [h!]
  \centering
	\includegraphics[width=0.45\textwidth]{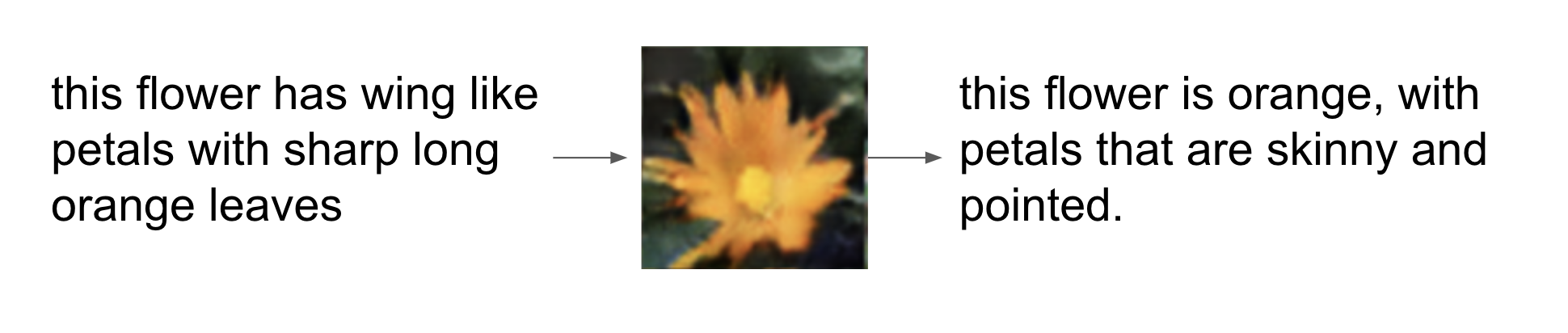}
  \caption{Example of going from text to image to text. Note that the text is different, but the meaning of the sentence is preserved.}
\end{figure}

\section{Conclusion}

We propose a novel way of generating pairs of image and text by repurposing the existing GAN-CLS architecture. This method can be extended for generating pairs of novel samples in multiple domains. Additionally, we study the implication of cycles, going from image to text to image and vise versa, as well as compare our model with autoencoders.

\end{document}